\documentclass[10pt, letterpaper, conference]{ieeeconf}
\IEEEoverridecommandlockouts                              

\usepackage[]{graphicx}
\usepackage[cmex10]{amsmath}
\usepackage{amsfonts, amssymb}
\usepackage{mdwmath}
\usepackage{mdwtab}
\usepackage[tight,footnotesize]{subfigure}
\usepackage{algorithm}    
\usepackage{algpseudocode}
\usepackage{balance}
\usepackage{comment}
\usepackage{gensymb}
\usepackage{hyperref}
\usepackage{multirow}

\let\oldhat\hat

\renewcommand{\hat}[1]{\oldhat{\mathbf{#1}}}

\usepackage{eso-pic}
\usepackage{xspace}
\makeatletter
\DeclareRobustCommand\onedot{\futurelet\@let@token\@onedot}
\def\@onedot{\ifx\@let@token.\else.\null\fi\xspace}

\def\eg{\emph{e.g}\onedot} 
\def\ie{\emph{i.e}\onedot}

\def\etal{\emph{et al}\onedot}
\makeatother

\title{\LARGE \bf
Discrete Rotation Equivariance for Point Cloud Recognition
}

\author{
Jiaxin Li$^{1}$, Yingcai Bi$^{2}$, Gim Hee Lee$^{1}$
\thanks{$^{1}$Jiaxin Li and Gim Hee Lee are with the Computer Vision and Robotic Perception (CVRP) Lab, School of Computing, National University of Singapore. \tt\small jli@u.nus.edu}%
\thanks{$^{2}$Yingcai Bi is with the Control Science Group, Temasek Laboratories, National University of Singapore.}%
\thanks{*Source codes at \url{https://github.com/lijx10/rot-equ-net}}
}

\begin{document}
\maketitle

\begin{abstract}
Despite the recent active research on processing point clouds with deep networks, few attention has been on the sensitivity of the networks to rotations. In this paper, we propose a deep learning architecture that achieves discrete $\mathbf{SO}(2)$/$\mathbf{SO}(3)$ rotation equivariance for point cloud recognition. Specifically, the rotation of an input point cloud with elements of a rotation group is similar to shuffling the feature vectors generated by our approach. The equivariance is easily reduced to invariance by eliminating the permutation with operations such as maximum or average. Our method can be directly applied to any existing point cloud based networks, resulting in significant improvements in their performance for rotated inputs. We show state-of-the-art results in the classification tasks with various datasets under both $\mathbf{SO}(2)$ and $\mathbf{SO}(3)$ rotations. In addition, we further analyze the necessary conditions of applying our approach to PointNet~\cite{qi2017pointnet} based networks.
\end{abstract}

\section{Introduction} 

Feature extraction is a fundamental problem in computer vision and it is the key to applications such as 
place recognition, Simultaneous Localization and Mapping (SLAM), object classification, etc. Techniques for feature extraction has evolved from handcrafted designs, e.g. SIFT \cite{ng2003sift}, ORB \cite{rublee2011orb}, to deep learning based methods, e.g. Convolutional Neural Networks (CNNs) \cite{krizhevsky2012imagenet}. All extraction methods share a common goal - to achieve equivariance to transformations of the input data that includes translation, scale, rotation. An informal definition of transformation equivariance is that the extracted features are predictable when a transformation is applied to the input data. Furthermore, invariance is a special case of equivariance where features are identical under transformation. Transformation equivariant / invariant features are necessary for robotics applications because sensors on robots such as UAVs or self-driving cars are constantly in motion. Objects and scenes are usually captured by the sensors at unknown poses but they should be classified in the correct categories or recognized as seen previously.

In the recent years, CNNs are shown to be more effective than hand-crafted methods in many tasks. CNNs are translational equivariant because the convolution kernel remains the same over the feature map. More specifically, shifting the input image and then applying the convolutional kernel is the same as applying convolutional kernels and then shifting the feature map. In addition, pooling enables invariance to local deformations \cite{lecun2015deep}. However, it is non-trivial to extend equivariance to more general transformations such as rotation. On the other hand, rotation is one of the most common transformations in robotics tasks such as loop closure detection and object detection. In this paper, we focus on solving the equivariance for $\mathbf{SO}(2)$/$\mathbf{SO}(3)$ rotations.

Data augmentation \cite{van2001art} is the most popular technique to mitigate the effects from rotations. Despite the simplicity, it often leads to larger amount of model parameters and is prone to under- or over-fitting \cite{laptev2016ti}. Another drawback of data augmentation is its black-box nature, where it is completely unknown on how the network handles various transformation. Systematic and explainable method by constraining the filter \cite{thomas2018tensor,worrall2017harmonic}, extending the convolutional domain to general groups \cite{cohen2016group,worrall2018cubenet,cohen2018spherical,esteves2018learning}, and making use of log-polar transforms \cite{esteves2017polar,henriques2016warped}, etc, have been proposed recently. 

Unfortunately, most studies on rotation equivariance focus on 2D convolution because the generalizations to 3D are difficult. Among the methods that work with 3D data, voxelization \cite{worrall2018cubenet,esteves2017polar,esteves2018learning} relies on 3D voxel grid which is computationally heavy and not scalable. Multi-view rendering \cite{su2015multi} is usually invariant to discrete 2D rotations but it suffers from information loss \cite{klokov2017escape} and few research has been done to study the effect of rotations in $\mathbf{SO}(3)$. On the contrary, point cloud is a compact representation of 3D objects and widely used by sensors like RGBD camera, LiDAR. PointNet \cite{qi2017pointnet} is the pioneer of consuming point clouds using deep networks, and many variants are later proposed base on it. Despite the use of data augmentation to handle rotated point clouds, few research has been done on rotation equivariance of point clouds.

We fill the gap by proposing an architecture to realize discrete $\mathbf{SO}(2)$/$\mathbf{SO}(3)$ equivariance for point cloud recognition. We define a $k$-element rotation group $R=\{r_i, i = [0, \cdots, k-1]\}$. For each input point cloud, we rotate it with each element in $R$ and feed it into the network to get a feature vector $f_i$. It can be proved that rotating the point clouds with $R$ equals to permuting the resulted $k$ feature vectors.

Our major contributions are as follows:
\begin{itemize}
    \item A general method that can be applied to existing networks to achieve discrete $\mathbf{SO}(2)$/$\mathbf{SO}(3)$ equivariance.
    \item We significantly improve the performance of processing rotated point clouds and achieve state-of-the-art performance on classifying ModelNet40/10, rotated MNIST.
    \item We analyze the necessary conditions of enhancing PointNet based networks with our method.
\end{itemize}

\section{Related Work} 
Rotation invariance is one of the desired properties when designing handcrafted features. SIFT \cite{ng2003sift} feature descriptor encodes orientation information by building a histogram to find the dominant gradient direction. ORB \cite{rublee2011orb} descriptor estimates an orientation of the local patch and rotates it to a canonical pose. For 3D point clouds, Normal Aligned Radial Feature (NARF) \cite{steder2010narf} and Rotation-Invariant Feature Transform (RIFT) \cite{lazebnik2004semi} are rotation invariant. NARF is invariant against rotation around the surface normal by calculating the dominant intensity change direction. RIFT builds a histogram according to neighbors' distances and gradient angles. Unlike deep learning based features, handcrafted features are not adaptive to tasks and hence have poor performance in recognition tasks like classification.

As mentioned, classical CNNs or MLPs are not equivariant to rotation. Steerability is first introduced by Teo \etal \cite{jacobsen2017dynamic} and recently used to put constraints on the network weights and architectures \cite{cohen2016steerable,kondor2018generalization,kondor2018n,cohen2018intertwiners}. Similarly, Worrall \etal \cite{worrall2017harmonic} present Harmonic Networks that achieve equivariance to patch-wise $360^{o}$-rotation by replacing CNN filters with circular harmonics. However, it is non-trivial to extend these 2D filter constraints to 3D inputs like voxel grids and point clouds. Recently the capsule network \cite{sabour2017dynamic,hinton2018matrix} is proposed to learn approximations to transformation equivariance, while few research has been done to explore its potential on point clouds.

Spatial Transformer Network (STN) \cite{jaderberg2015spatial} attempts to learn the parameters of the transformation given the structure. Though not theoretically equivariant, STNs perform well in image and 3D applications. Laptev \etal. \cite{laptev2016ti} realize scale and rotation invariance by pooling feature vectors over the input orbit, while it is done only with images and unclear how well it generalizes to point clouds. MVCNN \cite{su2015multi} shares similar idea but inherently it is based on image convolution and lack of study on $\mathbf{SO}(3)$ rotation. Cohen and Welling \cite{cohen2016group} extend the domain of 2D CNNs from translation to general groups. The idea of group equivariance is further extended to 3D voxel grids by CubeNet \cite{worrall2018cubenet}. Cohen \etal \cite{cohen2018spherical} combine this idea with generalized Fourier transforms to achieve rotation equivariance on $\mathbf{SO}(3)$. Polar Transformer Network (PTN) \cite{esteves2017polar} applies STN to predict an origin or axis and then projects images or voxels to the polar coordinate representation. Another drawback of PTN is the difficulty to achieves full $\mathbf{SO}(3)$ rotation equivariance. Thomas \etal propose the tensor field network \cite{thomas2018tensor}, which to the best of our knowledge is the first to achieve $\mathbf{SO}(3)$ rotation, translation and permutation of 3D points. The filters are constrained to be the product of spherical harmonic radial functions.

Following Cohen \& Welling \cite{cohen2016group}, we apply the idea of group equivariance to existing point cloud-based networks. Discrete rotation equivariance is achieved by training networks within a pre-defined rotation group. With intensive experiments, we show the effectiveness of our proposed architecture and the necessary conditions of applying it to point clouds.

\section{Rotation Equivariance Architecture} 
In this section, we present the definition and application of rotation group in Section~\ref{sec_group_def} and \ref{sec_3d_rot_group}. Based on the definition and application, we design and prove an equivatiant learning architecture in Section~\ref{sec_design}.

\subsection{Definition of Group} \label{sec_group_def}
Given a set $G$, and an operation $\circ$ that combines any two elements of $G$, \eg, $a, b$, the combination is denoted as $a\circ b$ or simply $ab$. Four properties are required to make $(G, \circ)$ a valid group.
\subsubsection{Closure} $\forall a,b \in G, ab \in G$
\subsubsection{Associativity} $\forall a,b,c \in G, (ab)c=a(bc)$
\subsubsection{Identity element} $\exists e\in G, ae=ea=a$
\subsubsection{Inverse element} $\forall a\in G, \exists b\in G, s.t., ab=ba=e$

We can easily construct a 2D rotation group $R=\{r_i=\frac{2\pi i}{k}, i=0,1,\cdots,k-1\}$. However, a group may not satisfy \textit{commutativity}, \ie, $ab$ is not necessarily $ba$. $\mathbf{SO}(2)$ rotation groups satisfy commutativity while $\mathbf{SO}(3)$ groups do not, which makes it more difficult to achieve $\mathbf{SO}(3)$ equivariance \cite{thomas2018tensor,aja2009tensors}.

\subsection{$\mathbf{SO}(3)$ rotation Group} \label{sec_3d_rot_group}
Besides the commutativity problem, it is not as obvious as the 2D case to construct a $\mathbf{SO}(3)$ group. Here we introduce the cube group and 2 other simplified rotation groups, i.e., Tetrahedral 12-group and Klein's 4-group.

\begin{figure}[t!]
        \centering
        \subfigure[]{\includegraphics[width=0.15\textwidth]{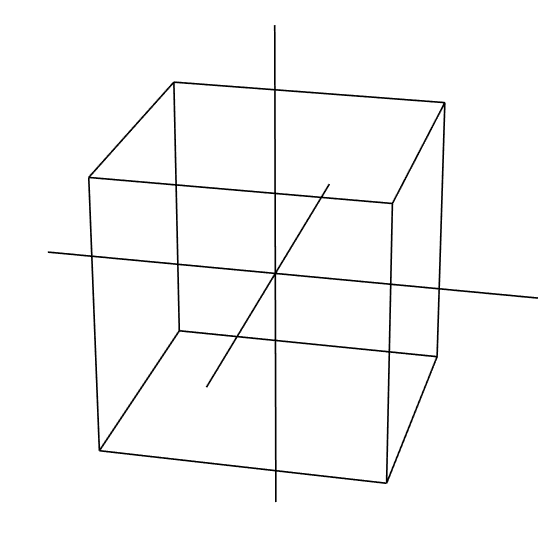}}
        \subfigure[]{\includegraphics[width=0.15\textwidth]{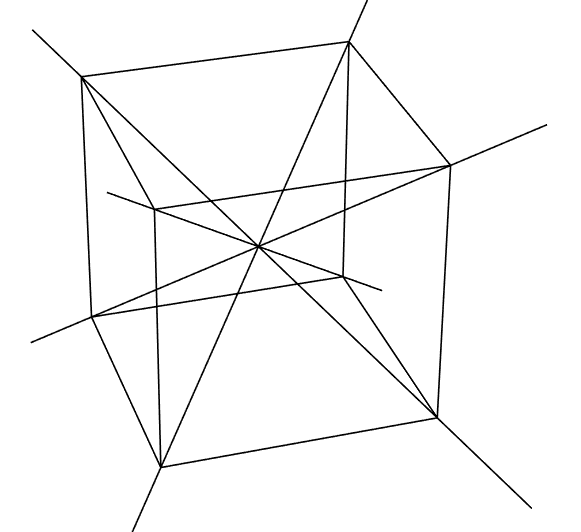}}
        \subfigure[]{\includegraphics[width=0.15\textwidth]{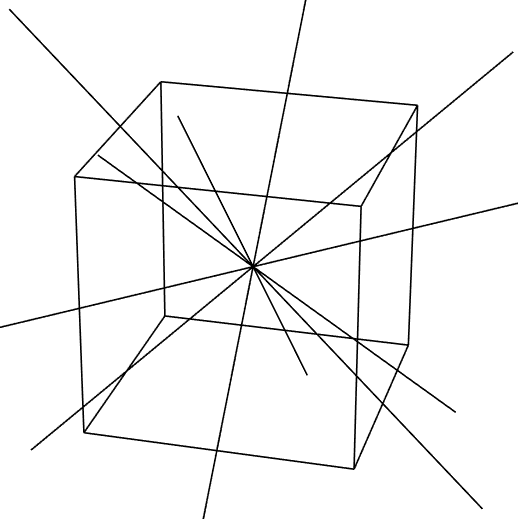}}
        \caption{Cube group of 24 rotations. (a) 90\degree, 180\degree and 270\degree rotation around the opposite faces. (b) $\pm120\degree$ rotation around the 4 diagonal axes. (c) 180\degree rotation about the 6 pairs of opposite edges. } \label{fig_cube_group}
        \vspace{-8pt}
\end{figure}

\begin{figure*}[th!] \centering
\includegraphics[width=0.98\textwidth]{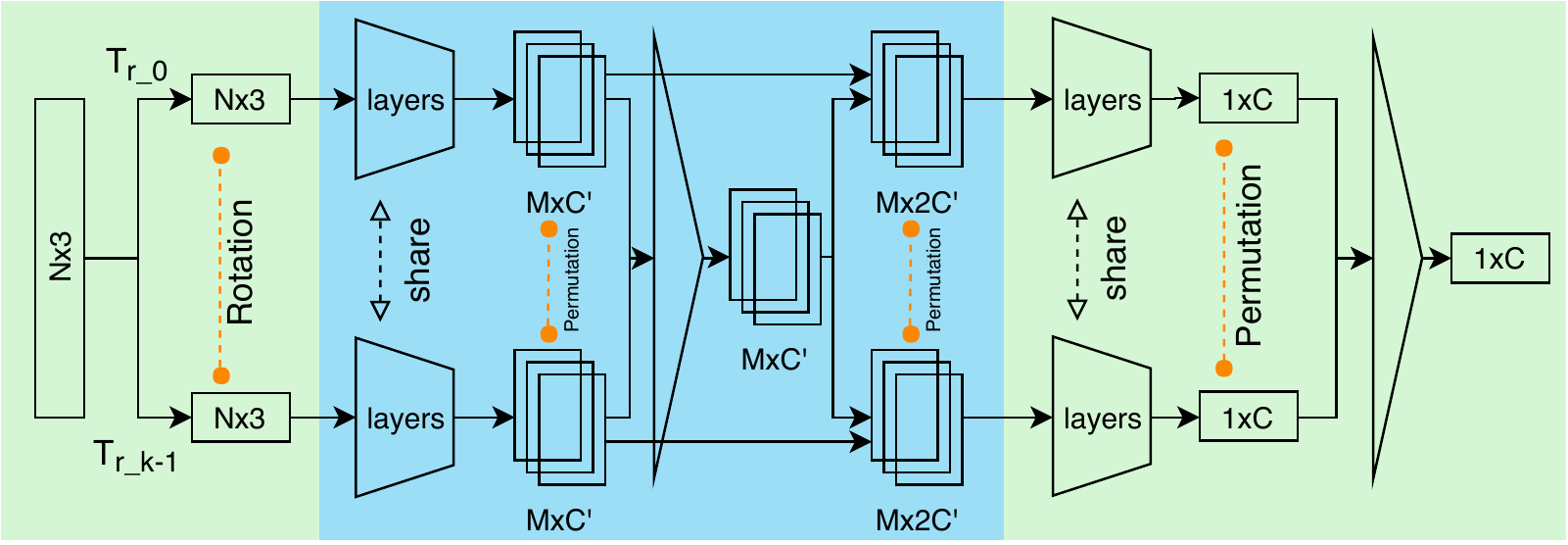}
\caption{The architecture of our rotation equivariant method. The green part (left \& right) shows the basic structure. Each input point cloud is rotated with every element $r_i$ in a rotation group $R$, and fed into a shared network. The blue (middle) part integrates this structure in a hierarchical network like SO-Net \cite{li2018so}. The rotation of input equals to the permutation of the features. }\label{fig_architecture}
\vspace{-8pt}
\end{figure*}

\subsubsection{Cube Group}
A cube has 24 rotation symmetry elements, called the cube group $S_4$. The cube group is the combination of any $\pm90\degree, \pm180\degree$ rotations around the three axes. Another intuitive interpretation of the cube group is shown in Fig.~\ref{fig_cube_group}. It consists of: (a) 90\degree, 180\degree and 270\degree rotation around the opposite faces. (b) $\pm120\degree$ rotation around the 4 diagonal axes. (c) 180\degree rotation about the 6 pairs of opposite edges. (d) Identity.

Our proposed approach requires feature pooling from vectors generated by each element in the rotation group. Smaller rotation groups require less computational power. In fact, our results show that considering a small rotation group already brings significant improvement, while enlarging the group brings marginal improvement.

\subsubsection{Tetrahedral 12-group}
As a subset of the cube group, the 12-group is the combination of any $\pm90\degree$ rotations around the three axes of a cube. 

\subsubsection{Klein's 4-group}
The Klein's 4-group is the smallest non-cyclic rotation group, which is the combination of any $180\degree$ rotations around the three face-to-face axes of a cube.

\subsection{Equivariant Design} \label{sec_design}

As mentioned above, an equivariant network should give predictable output when some transformation is applied to the input. Formally, given a network or a layer $\Phi$ and an input data $x$, transformation (in our case rotation) group $R=\{r_i\}$ and operator $T$, the equivariance requires,
\begin{equation} \label{equ_equivariance}
    \Phi(T_{r_i}x) = T'_{r_i}\Phi(x)
\end{equation}
where $T'_{r_i}$ is not necessarily the same as $T_{r_i}$. For instance, equivariance becomes invariance if $T'_{r_i}$ is identity $T_I$. In our method, $T'_{r_i}$ is a permutation.

We propose a network architecture that is equivariant to a discrete rotation group $R=\{r_i, i=0,\cdots,k-1\}$, which is shown in Fig.~\ref{fig_architecture}. The green part of Fig.~\ref{fig_architecture} is the basic structure of our method. For every input data $x$, we rotate it with every element in the group $R$, and feed it into a shared network to obtain $k$ feature vectors. Rotating $x$ with any element in $R$ equals to permuting the $k$ vectors, because of the closure property of the group. $R$ can be a simple 2D rotation group or a $\mathbf{SO}(3)$ group discussed in Section~\ref{sec_3d_rot_group}. Theoretical proof of the equivariance is presented below.

In applications like classification, the permutation is easily converted to invariance with operations like max-pooling, average-pooling, etc.

\subsubsection{Proof of Equivariance} \label{sec_proof}
In the context of point cloud, we denote a point cloud as $x\in \mathbb{R}^{N\times 3}$, which stores the 3D coordinates of $N$ points. A point cloud is orderless, in another word, the rows of $x$ can be permuted arbitrarily. Therefore a point cloud cannot be simply represented as a function $f:\mathbb{Z}^2/\mathbb{Z}^3 \to \mathbb{R}^C$, which is widely used in images \cite{cohen2016group} or voxel grids \cite{worrall2018cubenet}. Here we denote a rotation group as $R=\{r_i\in \mathbb{R}^{3\times 3}, i=0,1,\cdots,k-1\}$ and a rotation transform is defined as,
\begin{equation}
    T_{r_i}x = (rx^T)^T=xr^T
\end{equation}
therefore, operator $T$ is linear with respect to $R$,
\begin{equation}
    T_{r_i}T_{r_j}x = x r^T_j r^T_i = x(r_i r_j)^T = T_{r_i r_j}x
\end{equation}

Our network structure $\Phi$ is denoted as Eq.~\ref{eq_equivariance_left}, where $\phi$ represents the shared network or layer. It is visualized at the green part of Fig.~\ref{fig_architecture}.
\begin{equation} \label{eq_equivariance_left}
    \Phi(x) = [\phi(T_{r_0}x), \phi(T_{r_1}x), \cdots, \phi(T_{r_{k-1}}x)]^T
\end{equation}

Each element in group $R$ is unique, namely $\forall i\neq j, T_{r_i}\neq T_{r_j}$. Obviously, combining two different elements with a common rotation $T_r$ results in two different rotations.
\begin{equation} \label{eq_unique}
    T_{r_i}T_r=T_{r_i r} \neq T_{r_j}T_r=T_{r_j r}
\end{equation}

Therefore, each element in Eq.~\ref{eq_equivariance_right} is unique. Note that we enumerate all group elements in Eq.~\ref{eq_equivariance_left} and a group is close. Therefore given $r\in R$, Eq.~\ref{eq_equivariance_left} and \ref{eq_equivariance_right} share the same content except that the row ordering is different.
\begin{equation} \label{eq_equivariance_right}
    \Phi(T_r x) = [\phi(T_{r_0}T_r x), \phi(T_{r_1}T_r x), \cdots, \phi(T_{r_{k-1}}T_r x)]^T
\end{equation}

In summary, our design is rotation equivariant with respect to $R$, because rotating input $x$ equals to permuting the output of our network.
\begin{equation}
    \forall r\in{R}, \exists P, s.t. \Phi(T_r x) = P \Phi(x)
\end{equation}

\subsubsection{Integration with Hierarchical Networks} \label{sec_design_integration}
Theoretically, our approach can be applied to any network by rotating the input and then obtaining a set of permuted features. However, we find that at least the vanilla PointNet \cite{qi2017pointnet} is not suitable for such equivariant design. Instead, a hierarchical design is necessary for PointNet based networks. Details are given in Section~\ref{exp_pointnet}.

PointNet \cite{qi2017pointnet} is the pioneer of processing orderless points directly, but there is only one single global feature aggregation. Subsequent research like PointNet++ \cite{qi2017pointnet++}, SpiderCNN\cite{xu2018spidercnn}, SO-Net\cite{li2018so} point out that encoding multiple layers of local information increase the performance, similar to the ConvNet-like hierarchical feature aggregation. These methods achieve state-of-the-art performance and still maintain invariant to point permutation. In this paper we call this category of approaches as the hierarchical networks.

Here we take the example of SO-Net to demonstrate that our equivariant design can be integrated into each hierarchy of a network, to obtain both local and global equivariant features. The same method can be easily adapted to other hierarchical PointNet based networks. SO-Net takes a $N\times 3$ point cloud and produces a feature map of size $M\times C'$, which maintains the spatial distribution of points, encodes local information and is invariant to the order of points. The $M\times C'$ feature map can be further processed into another feature map of the same or different size.

As shown in the middle part of Fig.~\ref{fig_architecture}, we obtain $k$ $M\times C'$ feature maps on each level of SO-Net. A rotation invariant feature is built by pooling from these $k$ feature maps, and later concatenated back to the $k$ feature maps. The resulted $k$ $M\times 2C'$ feature maps are still rotation equivariant. These local feature maps are useful for applications that require local and global feature integration like per-point semantic segmentation.

\section{Experiments} 
\begin{table*}[th!]
\centering
\caption{Classification results on ModelNet40/10 under z rotations.}
\label{tbl_exp_sonet}
\begin{tabular}{l|l|ll|l||l|l|ll|l}
\hline
\multirow{2}{*}{Method} & \multirow{2}{*}{Representation} & \multicolumn{2}{l|}{ModelNet40} & \# of params & \multirow{2}{*}{Method} & \multirow{2}{*}{Representation} & \multicolumn{2}{l|}{ModelNet10} & \# of params \\
                        &                                 & Ins.           & Cls.           & $\times 10^6$        &                         &                                 & Ins.           & Cls            & $\times 10^6$        \\ \hline
PointNet\cite{qi2017pointnet}                & 1024 points                     & 89.2           & 86.2           & 3.5          & 3D ShapeNets\cite{wu20153d}            & $30^3$ voxel                      & 83.5           & -              & 12           \\
PointNet++\cite{qi2017pointnet++}              & 5000 points                     & 91.9           & -              & 1.1          & VRN\cite{brock2016generative}                     & $32^3$ voxel                      & 91.3           & -              & 18           \\
DeepSets\cite{zaheer2017deep}                & 5000 points                     & 90.0           & -              & -            & VoxNet\cite{maturana2015voxnet}                  & $30^3$ voxel                      & 92.0           & -              & 0.92         \\
SpiderCNN\cite{xu2018spidercnn}                     & 1024 points                     & 92.4           & -           & -            & Fusion-Net\cite{hegde2016fusionnet}              & $30^3$ voxel                      & 93.1           & -              & 120          \\
OctNet\cite{riegler2017octnet}                  & $128^3$ octree                    & 86.5           & 83.8           & -            & ORION\cite{sedaghat2016orientation}                   & $28^3$ voxel                      & 93.8           & -              & 0.91         \\
O-CNN\cite{wang2017cnn}                   & $64^3$ octree                     & 90.6           & -              & -            & CubeNet\cite{worrall2018cubenet}                 & $32^3$ voxel                      & \textbf{94.6}           & -              & 4.5          \\
PointCNN\cite{li2018pointcnn}                & 1024 points                     & 91.7           & -              & 0.45         & OctNet\cite{riegler2017octnet}                  & 1$28^3$ octree                    & 90.9           & 90.1           & -            \\
PTN\cite{esteves2017polar}                     & voxel                           & 89.9           & 86.5           & -            & ECC\cite{simonovsky2017dynamic}                     & 1000 points                     & 90.8           & 90.0           & -            \\
SO-Net\cite{li2018so}                  & 5000 points                     & 92.1           & 89.6           & 2.5          & SO-Net\cite{li2018so}                  & 5000 points                     & 93.9              & 93.9              & 2.5          \\ \hline
Rot-SO-Net - 1 rot            & 1024 points                     & 91.7           & 89.4           & 2.5          & Rot-SO-Net - 1 rot            & 1024 points                     & 92.5           & 92.1           & 2.5          \\
Rot-SO-Net - 4 rot            & 1024 points                     & 92.1           & 89.7           & 2.5          & Rot-SO-Net - 4 rot            & 1024 points                     & 93.7           & 93.6           & 2.5          \\
Rot-SO-Net - 6 rot            & 1024 points                     & 92.3           & 90.2           & 2.5          & Rot-SO-Net - 6 rot            & 1024 points                     & 94.3           & 94.3           & 2.5          \\
Rot-SO-Net - 9 rot            & 1024 points                     & 92.4           & 90.2           & 2.5          & Rot-SO-Net - 9 rot            & 1024 points                     & 94.3           & 94.3           & 2.5          \\
Rot-SO-Net - 12 rot           & 1024 points                     & \textbf{92.5}           & \textbf{90.4}           & 2.5          & Rot-SO-Net - 12 rot           & 1024 points                     & 94.5           & 94.5           & 2.5          \\ \hline
\end{tabular}
\vspace{-8pt}
\end{table*}
\begin{table}[t]
\centering
\caption{Classification results on MN40 under $\mathbf{SO}(3)$/z rotations.}
\label{tbl_exp_sonet_3d}
\begin{tabular}{l|l|l|l|l}
\hline
Method         & Input size     & param \# & $\mathbf{SO}(3)$ & z \\ \hline
PointNet\cite{qi2017pointnet}       & $2048\times 3$    & 3.5M        & 83.6 & 89.2    \\
VoxNet\cite{maturana2015voxnet}         & $30^3$     & 0.9M        & 73.0 & 83.0    \\
SubVolSup\cite{qi2016volumetric}      & $30^3$     & 17M         & 82.7 & 88.5    \\
SubVolSup MO\cite{qi2016volumetric}   & $20\times 30^3$  & 17M         & 85.0 & 89.5    \\
MVCNN 12x\cite{su2015multi}      & $12\times 224^2$ & 99M         & 77.6 & 89.5    \\
MVCNN 80x\cite{su2015multi}      & $80\times 224^2$ & 99M         & 86.0 & 90.2    \\
Spherical CNNs\cite{esteves2018learning} & $2\times 64^2$   & 0.5M        & 86.9 & 88.9    \\ \hline
Rot-SO-Net - 1 rot   & $1024\times 3$    & 2.5M        & 84.4 & 91.7 \\
Rot-SO-Net - 4 rot   & $1024\times 3$    & 2.5M        & 87.7 & 92.1 \\
Rot-SO-Net - 12 rot  & $1024\times 3$    & 2.5M        & \textbf{88.8} & \textbf{92.5} \\ \hline
\end{tabular}
\end{table}
\begin{table}[th!]
\centering
\caption{Classification result on rotated MNIST dataset.}
\label{tbl_exp_mnist}
\begin{tabular}{l|l|l}
\hline
Method                                         & Error (\%) & \# of param $\times 10^6$ \\ \hline
TI-Pooling\cite{laptev2016ti}                  & 1.2        & $\approx 1$              \\
Polar Transformer Net\cite{esteves2017polar}   & \textbf{0.89}       & 0.25              \\
Group Equivariant CNN\cite{cohen2016group}     & 2.28       & 0.022             \\
Harmonic Networks\cite{worrall2017harmonic}    & 1.69       & 0.033             \\
RotEqNet\cite{marcos2017rotation}              & 1.01       & 0.1               \\
CNN\cite{cohen2016group}                       & 5.03       & 0.022             \\
ORN\cite{zhou2017oriented}                     & 2.25       & $\approx 1$               \\ \hline
Rot-SO-Net - 1 rot           & 5.57       & 1.45              \\
Rot-SO-Net - 4 rot           & 2.94       & 1.45              \\
Rot-SO-Net - 6 rot           & 2.72       & 1.45              \\
Rot-SO-Net - 9 rot           & 2.47       & 1.45              \\
Rot-SO-Net - 12 rot          & 2.22       & 1.45              \\ \hline
\end{tabular}
\vspace{-8pt}
\end{table}

In this section, we evaluate the effectiveness of our approach with classification on 3D point cloud datasets, namely ModelNet40/10 \cite{wu20153d}, and a 2D dataset rotated MNIST \cite{larochelle2007empirical}. For ModelNet40/10, we out-perform state-of-the-art methods under $\mathbf{SO}(2)$ and $\mathbf{SO}(3)$ rotations. For rotated MNIST, we demonstrate competitive accuracy even though we are the only approach that takes orderless pixel coordinates (2D point cloud) as input, while all others consume images as input and apply convolutions.

Our experiments consist of two parts and three types of point cloud networks. The first part is done with a recently proposed hierarchical network, namely SO-Net \cite{li2018so}, to demonstrate our best performance. The second part is done with a vanilla version and a simple variant of PointNet \cite{qi2017pointnet}, to explore the properties of our proposed architecture.

\subsection{Implementation} \label{sec_exp_implement}
All networks used in this paper are implemented with PyTorch on a Nvidia GTX1080ti GPU. We follow the default network structures and training configurations of SO-Net and PointNet.

\subsubsection{z Rotation} \label{sec_exp_implement_2d}
We apply random 2D rotation augmentation during training. For ModelNet40/10, the rotation is around the up-axis (z-axis), to simulate that the gravity direction is given while the horizontal orientation is unknown. When testing the ModelNet40/10, each point cloud is rotated 12 times to evenly cover $2\pi$ around the z-axis. The final classification result is obtained from the average score of the 12 outputs.

The rotated MNIST is itself randomly rotated, hence it is not necessary to do the averaging during testing.

\subsubsection{$\mathbf{SO}(3)$ Rotation} \label{sec_exp_implement_3d}
Similarly with Section~\ref{sec_exp_implement_2d}, we use random $\mathbf{SO}(3)$ rotation augmentation during training. When testing the ModelNet with $\mathbf{SO}(3)$ rotation, we randomly rotate each point cloud 12 times and employ the same averaging technique.

Since rotated MNIST is actually an image dataset, we do not perform $\mathbf{SO}(3)$ rotation on it.

\subsection{Datasets}
\begin{figure}[t] \centering
\includegraphics[width=0.45\textwidth]{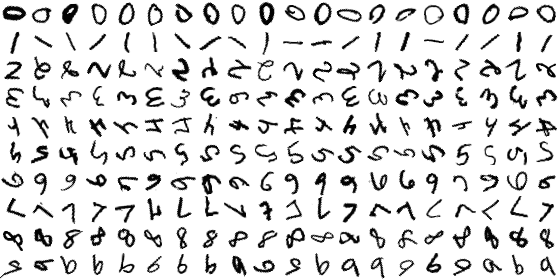}
\caption{Examples of handwriting digits from rotated MNIST.}\label{fig_exp_mnist_example}
\vspace{-8pt}
\end{figure}
\begin{figure}[t]
        \centering
        \includegraphics[width=0.11\textwidth]{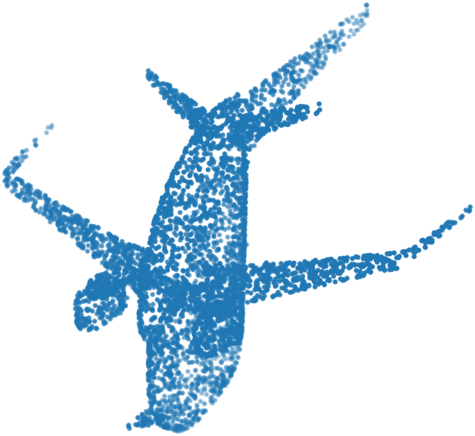}
        \includegraphics[width=0.11\textwidth]{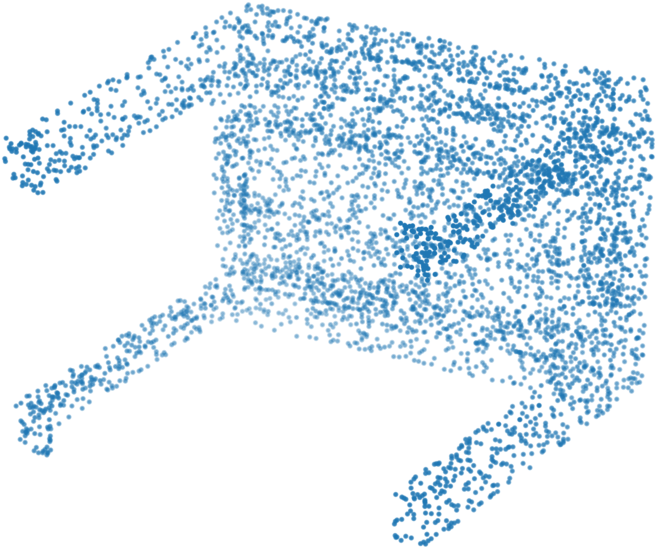}
        \includegraphics[width=0.11\textwidth]{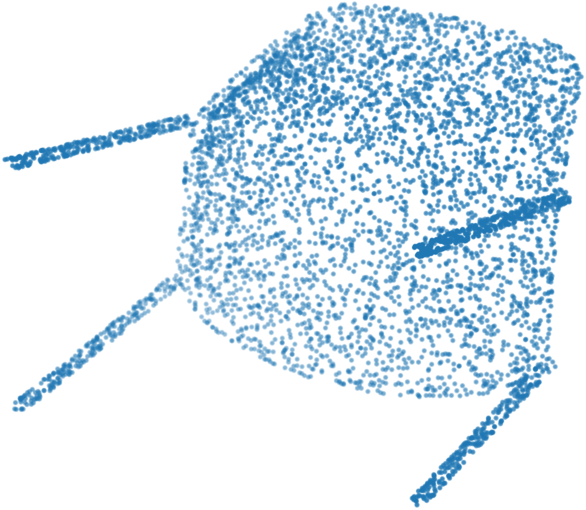}
        \includegraphics[width=0.11\textwidth]{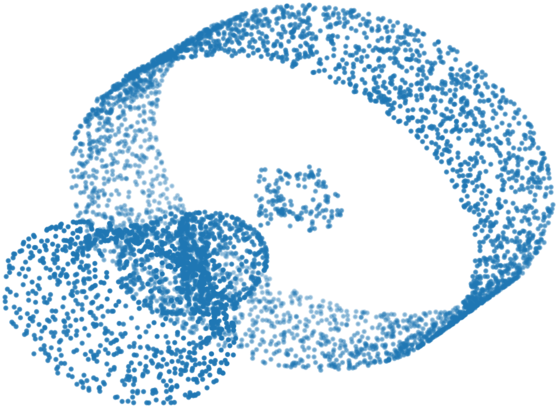}
        \caption{Examples of point clouds sampled from ModelNet dataset.}
        \label{fig_exp_modelnet_example}
        \vspace{-8pt}
\end{figure}

We use rotated MNIST to make comparison with existing rotation equivariant methods, because most research on equivariance is done on images. Rotated MNIST \cite{larochelle2007empirical} is a variant of the original MNIST \cite{lecun1998gradient} dataset, with 10,000 / 2,000 / 50,000 handwriting digit images for training, validation and testing, respectively. Each image is rotated randomly. The random rotation and the small size of training set make it much more difficult than the original MNIST. Examples from rotated MNIST are shown in Fig.~\ref{fig_exp_mnist_example}.

We take the non-zero pixel coordinates to build a 2D point cloud of size $N \times 2$. In addition, the intensity on each non-zero pixel is used as extra information and concatenated with the point coordinate to form an $N \times 3$ matrix. Linear upsampling is performed to get a $512\times 3$ point cloud from each image.

ModelNet40 contains 13,834 objects from 40 categories like desk, chair, etc. It is split into training and testing sets of size 9,843 and 3,991, respectively. Similarly, ModelNet10 contains 3,377 objects from 10 categories, and split into 2,468 training samples and 909 testing samples. Originally ModelNet is provided as mesh models, so we perform uniform sampling on each mesh to obtain 1024 points and the corresponding surface normal vectors. Here each point cloud is represented as a $1024\times 6$ matrix. We employ the prepared dataset from SO-Net \cite{li2018so}. Examples from ModelNet are shown in Fig.~\ref{fig_exp_modelnet_example}.

In terms of data augmentation, we follow the standard practice including random jittering with Gaussian noise, random scaling, etc.

\subsection{Hierarchical Networks} \label{sec_exp_hier}
In this section, we demonstrate the effectiveness of by integrating our approach with SO-Net \cite{li2018so}, called Rot-SO-Net. Experiments are performed in two configurations, namely z rotations and $\mathbf{SO}(3)$ rotations, as stated in Section~\ref{sec_exp_implement}.

\subsubsection{z Rotations} \label{sec_exp_hier_2d}
We define five z rotation groups with 1, 4, 6, 9, 12 elements, respectively, by uniformly dividing $2\pi$. For example, a 4-element z rotation group contains rotations of $\{0\degree, 90\degree, 180\degree, 270\degree \}$. In particular, the 1-element group is simply identity $\{0\degree\}$, which satisfies the four requirements defined in Section~\ref{sec_group_def}.

Comparison with existing algorithms is presented in Table~\ref{tbl_exp_sonet} with ModelNet40/10 dataset, where ``\# rot'' represents the number of elements in a rotation group. In ModelNet40/10, the gravity direction is given, \ie, the roll and pitch angles are zero. ModelNet10 is orientation-aligned, in another word, all objects are in a canonical horizontal orientation. Although ModelNet40 is theoretically not orientation-aligned, most objects in the dataset are actually aligned. Therefore applying random rotation augmentation, either z or $\mathbf{SO}(3)$ rotation, discards the known orientation and decreases the classification performance. Unfortunately, there isn't a consensus on whether rotation augmentation, usually z rotation augmentation, should be applied during training. In order to make a fair comparison, we only present methods that employ z rotation augmentation during training, as shown in Table~\ref{tbl_exp_sonet}.

We achieve state-of-the-art performance in both ModelNet40 and ModelNet10. In particular, our $92.5\%$ accuracy is slightly better than SpiderCNN \cite{xu2018spidercnn}, the current leader, on ModelNet40. Although we are 0.1\% behind CubeNet \cite{worrall2018cubenet} on ModelNet10, we have the advantage of smaller model size and much better scalability since we directly work on point clouds while CubeNet works on voxel grids. Our computation and memory cost is $O(N)$ with respect to the number of points, but voxel grid based networks are $O(N^3)$ with respect to the size of the grid.

Results on rotated MNIST is shown in Table~\ref{tbl_exp_mnist}. The application of our equivariant approach significantly improves the performance on rotated MNIST, from an error rate of 5.57\% to 2.22\%, which is lower than recently proposed methods like ORN \cite{zhou2017oriented} and Group Equivariant CNN \cite{cohen2016group}. Note that all methods except ours make use of ConvNets and work on images, while we directly work with point clouds and still achieve competitive results.

\subsubsection{$\mathbf{SO}(3)$ Rotations} \label{sec_exp_hier_3d}
Similarly, we define three $\mathbf{SO}(3)$ rotation groups of size 1, 4, 12 according to Section~\ref{sec_3d_rot_group}. Classification results on ModelNet40 are presented in Table~\ref{tbl_exp_sonet_3d}. Samples are rotated randomly in $\mathbf{SO}(3)$ during both training and testing. 

The $\mathbf{SO}(3)$ rotation significantly increases the difficulty of recognizing point clouds. All methods suffer a significant drop with $\mathbf{SO}(3)$ rotation. Our equivariant design reduces the $\mathbf{SO}(3)$ effect and outperforms state-of-the-art methods by 1.9\%. Even with a small rotation group of size 4, we still have an advantage of 0.8\% over Spherical CNNs \cite{esteves2018learning}.

\subsection{PointNet and its Variant} \label{exp_pointnet}
Given the effectiveness of our equivariant design shown above, we further explore whether our method is universally applicable to all point cloud based networks. In particular, we apply our approach to PointNet (PN) \cite{qi2017pointnet} and one of its variant called PointNet with feature encoding (PointNet-FE/PN-FE), which is proposed in VoxelNet \cite{zhou2017voxelnet}. Note that the PointNet used in this subsection is based on our own implementation without T-Net \cite{qi2017pointnet} of the original paper.

Instead of making use of local feature aggregation, PointNet-FE consists of two small PointNets to utilize the global feature information. The per-point features from the first PointNet are concatenated with the first global feature vector, and then fed into the second PointNet to acquire the second (final) global feature vector.

\begin{figure}[th!]
        \centering
        \subfigure[]{\label{fig_plot_mn40_2d}\includegraphics[width=0.23\textwidth]{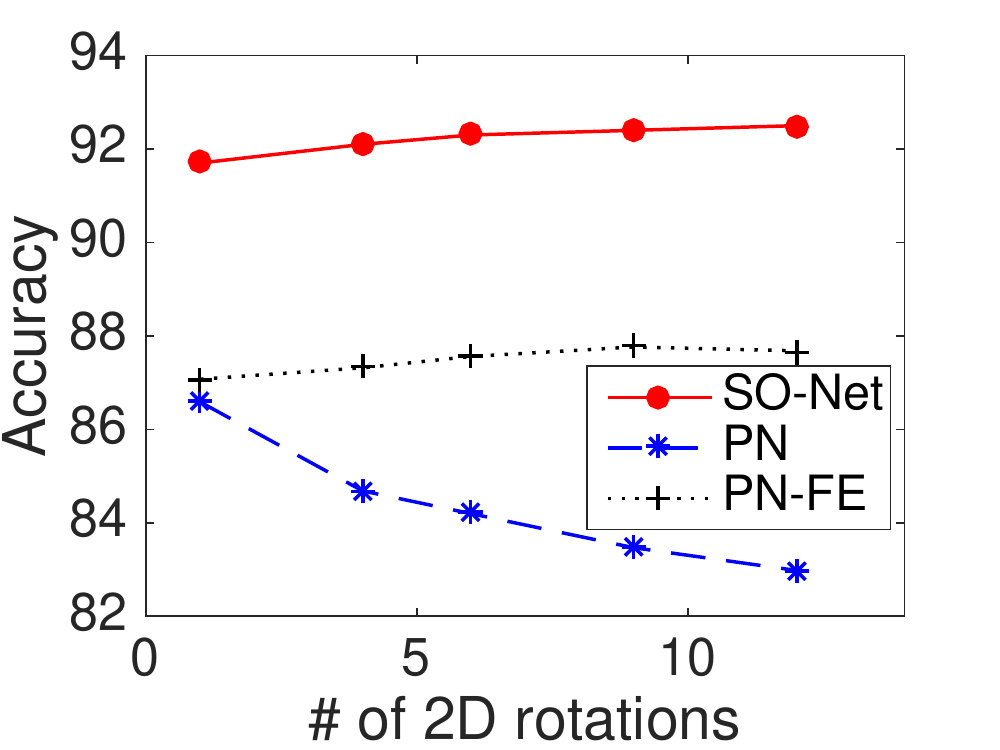}}
        \subfigure[]{\label{fig_plot_mn10_2d}\includegraphics[width=0.23\textwidth]{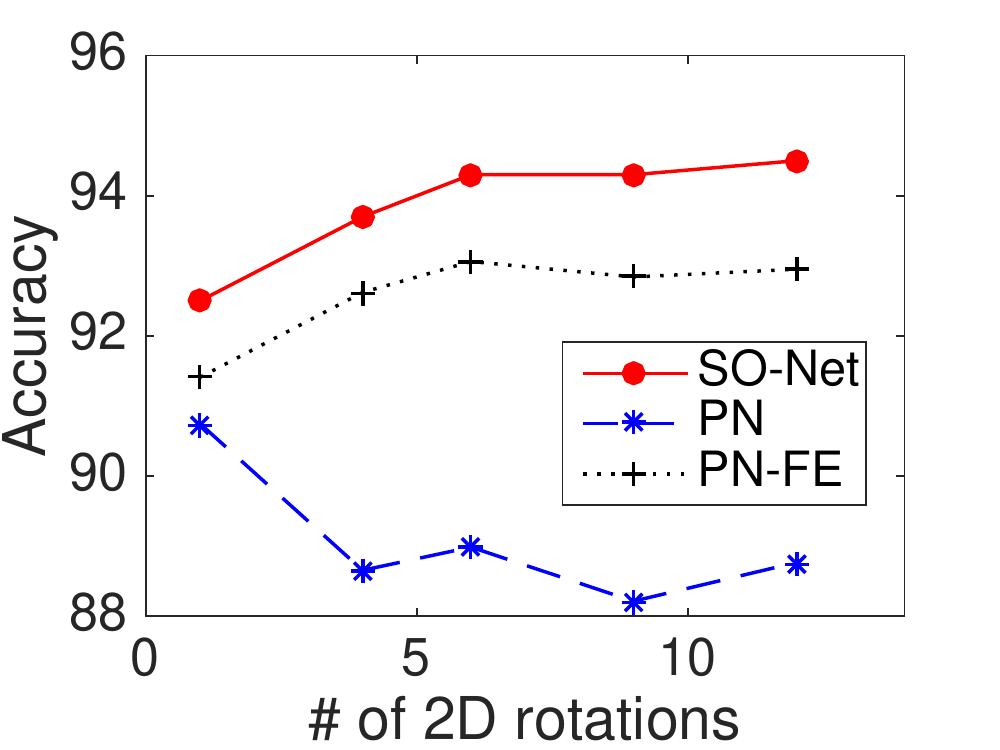}}
        \subfigure[]{\label{fig_plot_mn40_3d}\includegraphics[width=0.23\textwidth]{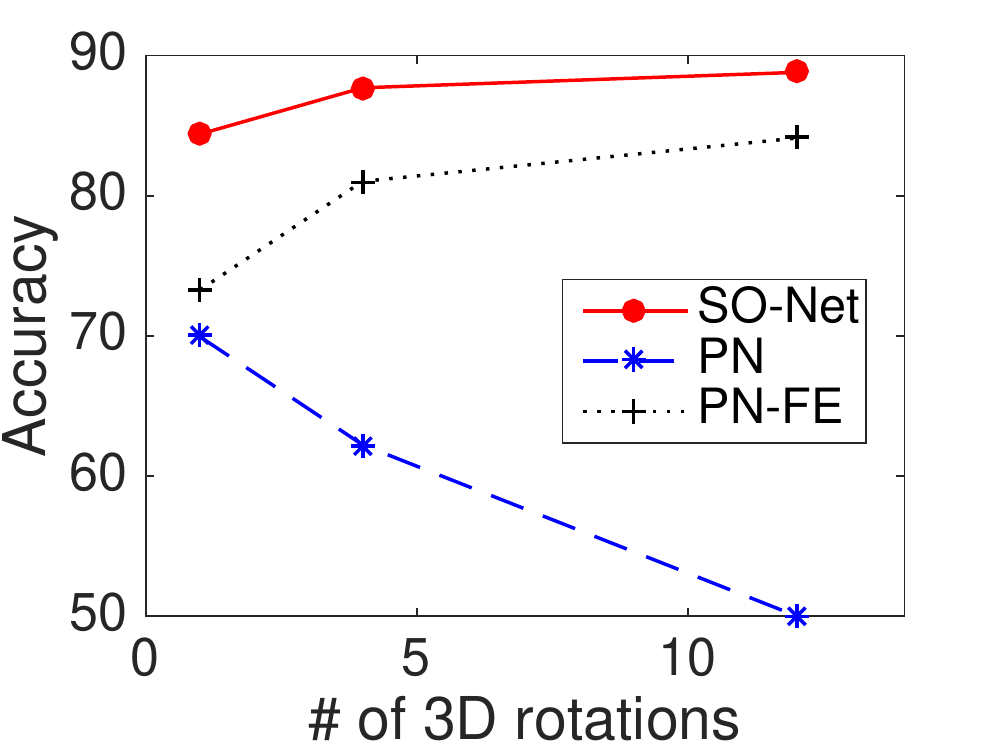}}
        \subfigure[]{\label{fig_plot_mn10_3d}\includegraphics[width=0.23\textwidth]{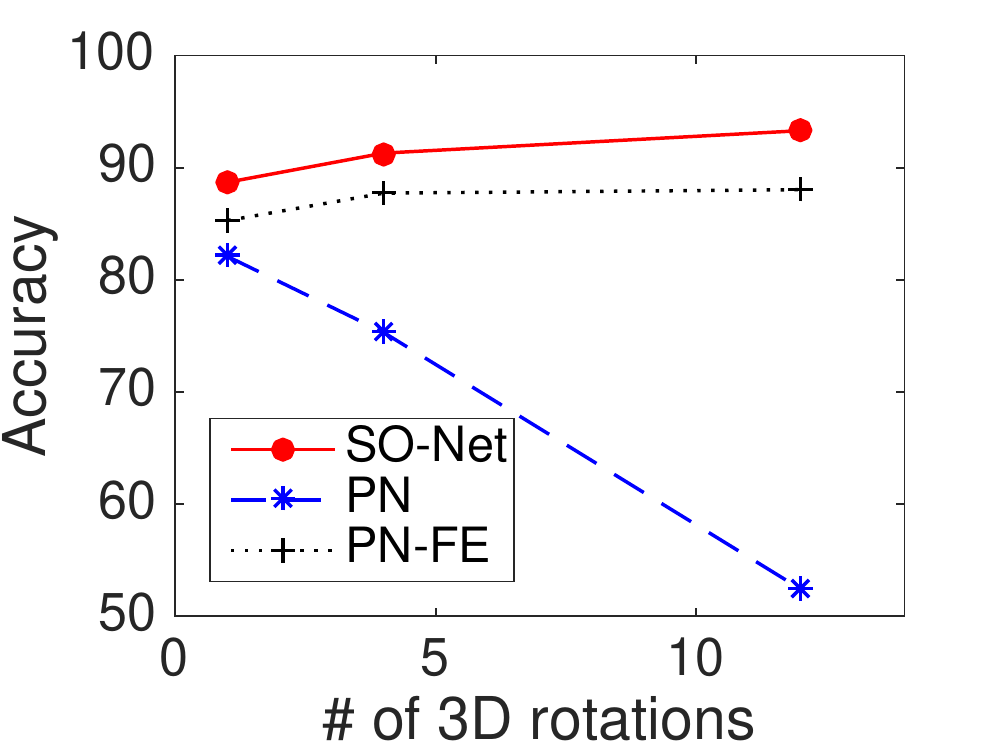}}
        \caption{Classification accuracy versus number of elements in the rotation group for (a) ModelNet40 under z rotations. (b) ModelNet10 under z rotations. (c) ModelNet40 under $\mathbf{SO}(3)$ rotations. (d) ModelNet10 under $\mathbf{SO}(3)$ rotations.}\label{fig_plot}
        \vspace{-8pt}
\end{figure}

As shown in Table~\ref{tbl_exp_pointnet_2d}, Table~\ref{tbl_exp_pointnet_3d_rot} and Fig.~\ref{fig_plot}, our method leads to significant accuracy improvement with PointNet-FE, espeically in 3D cases, similar to results shown in Section~\ref{sec_exp_hier}. Specifically, the accuracy on ModelNet40 and ModelNet10 significantly increases by 10.8\% and 2.7\% under $\mathbf{SO}(3)$ rotation, if the rot-12 rotation equivariant design is applied. On the contrary, the integration of PointNet and our method results in deteriorated performance. There is a performance drop of 19.9\% and 29.7\% under $\mathbf{SO}(3)$ rotation, after the application of rot-12 design.

Such contradictory results are clearly presented in Fig.~\ref{fig_plot}. Both SO-Net and PointNet-FE benefit from our equivariant design while PointNet shows negative results. Further experiments, though not illustrated in this paper, verify that increasing the depth or width does not make PointNet work under our architecture. On the other hand, smaller SO-Net or PointNet-FE still work well.

Both SO-Net and PointNet-FE share a common structure that there is more than one hierarchy, which serves to aggregate local and global information. On the other hand, the global-pooling design of vanilla PointNet does not allow information exchange between points. That is, each point or per-point feature is processed independently by a MLP. An intuitive guess is that, PointNet maps input points into a learned high dimensional space (what MLP is good at), without actively considering the neighboring or global-local relationship. With our design, the input points distribute more uniformly in the original 3D/2D space because each point cloud is rotated multiple times and fed into PointNet together. That is likely to make it more difficult to learn the mapping. Take an example that rotating two points around their center, the trajectory will form a sphere. It is impossible to tell that sphere is actually built by two points, if we do not make use of the relationship between the two points.

\begin{table}[t]
\centering
\caption{Results on MN40/10 for PN/PN-FE under z rotations.}
\label{tbl_exp_pointnet_2d}
\begin{tabular}{l||ll|ll||ll|ll}
\hline
\multirow{3}{*}{Rot \#} & \multicolumn{4}{c||}{PointNet}                         & \multicolumn{4}{c}{PointNet-FE}                     \\ \cline{2-9} 
                        & \multicolumn{2}{c|}{MN40} & \multicolumn{2}{c||}{MN10} & \multicolumn{2}{c|}{MN40} & \multicolumn{2}{c}{MN10} \\
                        & Ins.        & Cls.        & Ins.        & Cls.        & Ins.        & Cls.        & Ins.        & Cls.       \\ \hline
1                       & \textbf{86.6}        & \textbf{82.5}        & \textbf{90.7}        & \textbf{89.9}        & 87.1        & 83.4        & 91.4        & 90.9       \\
4                       & 84.7        & 80.7        & 88.7        & 87.3        & 87.3        & 83.5        & 92.6        & 92.3       \\
6                       & 84.2        & 80.3        & 89.0        & 87.9        & 87.6        & 83.9        & \textbf{93.1}        & \textbf{92.6}       \\
9                       & 83.5        & 80.0        & 88.2        & 86.7        & \textbf{87.8}        & \textbf{84.0}        & 92.8        & 92.5       \\
12                      & 83.0        & 79.3        & 88.8        & 87.5        & 87.7        & 84.0        & 93.0        & 92.3       \\ \hline
\end{tabular}
\vspace{-4pt}
\end{table}
\begin{table}[t]
\centering
\caption{Results on MN40/10 for PN/PN-FE under $\mathbf{SO}(3)$ rotations.}
\label{tbl_exp_pointnet_3d_rot}
\begin{tabular}{l||ll|ll||ll|ll}
\hline
\multirow{3}{*}{Rot \#} & \multicolumn{4}{c|}{PointNet}                                                                               & \multicolumn{4}{c}{PointNet-FE}                                                                            \\ \cline{2-9} 
                        & \multicolumn{2}{c|}{ModelNet40}                      & \multicolumn{2}{c|}{ModelNet10}                      & \multicolumn{2}{c|}{ModelNet40}                      & \multicolumn{2}{c}{ModelNet10}                      \\
                        & \multicolumn{1}{c}{Ins.} & \multicolumn{1}{c|}{Cls.} & \multicolumn{1}{c}{Ins.} & \multicolumn{1}{c|}{Cls.} & \multicolumn{1}{c}{Ins.} & \multicolumn{1}{c|}{Cls.} & \multicolumn{1}{c}{Ins.} & \multicolumn{1}{c}{Cls.} \\ \hline
1                       & \textbf{69.9}                     & \textbf{64.2}                      & \textbf{82.1}                     & \textbf{80.9}                      & 73.3                     & 69.2                      & 85.3                     & 84.4                      \\
4                       & 62.2                     & 56.2                      & 75.3                     & 73.2                      & 81.0                     & 75.9                      & 87.7                     & 86.9                      \\
12                      & 50.0                     & 44.5                      & 52.4                     & 49.7                      & \textbf{84.1}                     & \textbf{79.0}                      & \textbf{88.0}                     & \textbf{87.2}                      \\ \hline
\end{tabular}
\vspace{-8pt}
\end{table}

\subsection{Discussions}
As shown in Fig.~\ref{fig_plot}, our equivariant design enhances networks that have more than one hierarchy. For SO-Net with ModelNet40, we bring the improvement of 0.8\% and 4.4\%, respectively under z and $\mathbf{SO}(3)$ rotation. For PointNet-FE with ModelNet40, we achieve 0.8\% and 10.8\% improvement respectively. The improvement on $\mathbf{SO}(3)$ rotation is much more significant, probably because data augmentation works relatively well for the 2D case. Another observation is that generally increasing the size of the rotation group brings better performance, but the marginal return is diminishing, and the computational cost is roughly linear with the size of the group. A group of size 4 usually produces considerable improvement while the additional computational cost remains acceptable.

\section{Conclusion}
In this paper, we propose a deep learning architecture that achieves discrete $\mathbf{SO}(2)$/$\mathbf{SO}(3)$ rotation equivariance for point cloud based networks. For rotations within a pre-defined rotation group, rotating the input point cloud equals to permutating the feature maps. Besides providing theoretical proof, we demonstrate state-of-the-art performance in classification tasks. Further analysis shows that multi-hierarchy structure is necessary for PointNet based methods to enjoy the benefits brought by our design. One of the future directions is to utilize the polar coordinate to turn rotation equivariance to translation equivariance.

\noindent\textbf{Acknowledgment.} This work was partially supported by the Singapore MOE Tier 1 grant R-252-000-637-112.

%


\footnotesize
\bibliographystyle{IEEEtran}
\bibliography{ICRA2019_REF}




\end{document}